\documentclass[letterpaper, 10 pt, conference]{ieeeconf}  

\IEEEoverridecommandlockouts                              

\makeatletter
\let\NAT@parse\undefined

\def\@maketitle{%
  \newpage
  \null
  \begin{center}
    {\LARGE \bf \@title \par}
    \vskip 1em

    {\large \@author \par}
    \vskip 2em

    \includegraphics[width=0.92\textwidth]{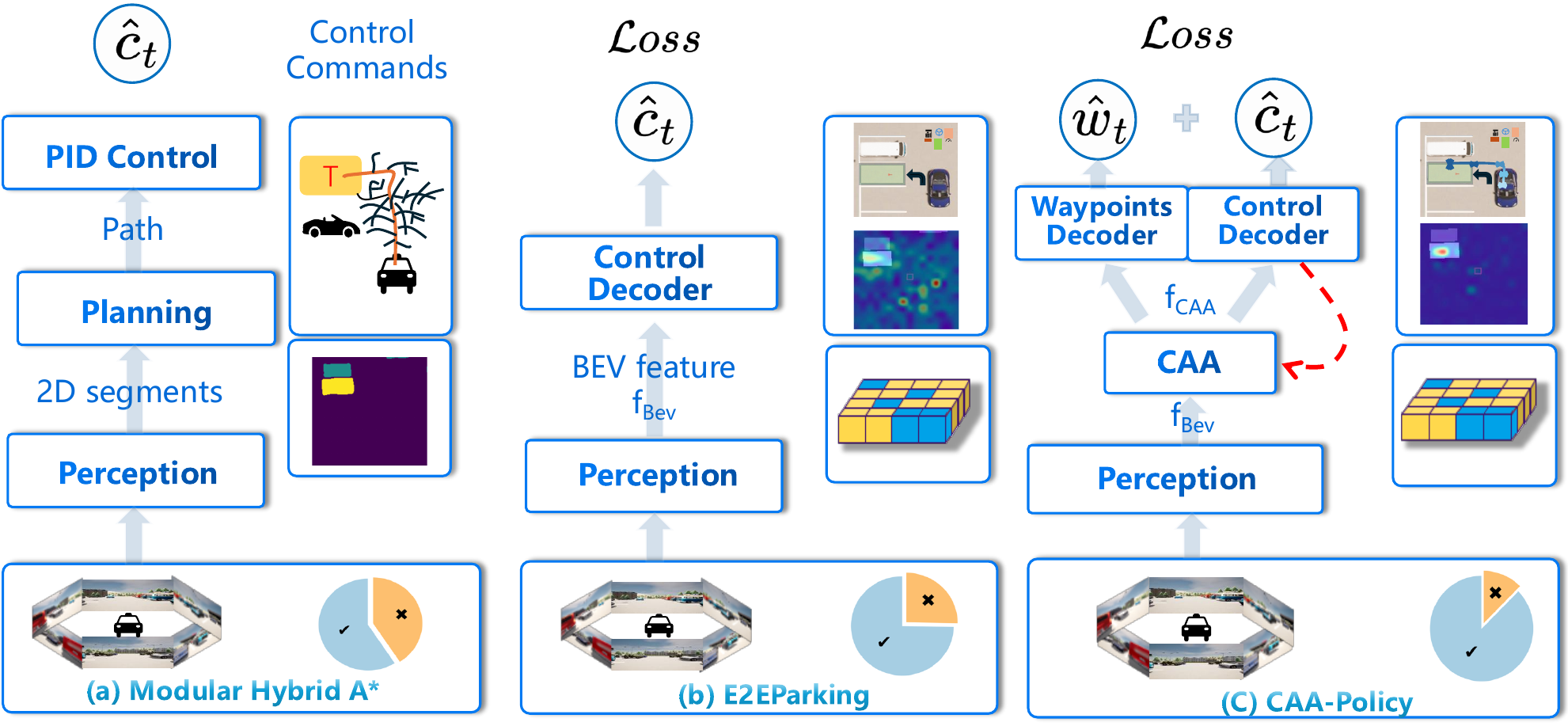}
    \vskip 0.5em
    {\small
    \stepcounter{figure}
    \textbf{Fig. \thefigure:} 
    Comparison of three paradigms, with perception outputs, planning paths, decoder attention, control/waypoints, and parking success rates (pie charts) shown. 
(a) Modular pipeline: learning-based perception produces BEV segmentation, followed by planning and PID control.
(b) E2EParking: end-to-end perception produces BEV features $\mathbf{f}_{\text{BEV}}$, decoded by a transformer to controls.
(c) CAA-Policy: CAA refines BEV features $\mathbf{f}_{\text{BEV}}$ into $\mathbf{f}_{\text{CAA}}$, where backpropagated gradients from control outputs (rather than solely from the training loss) adapt perception features. In addition, a waypoint decoder is introduced to preserve temporal consistency alongside control prediction.

    }
    \label{fig:teasing} 
    \vspace{-6mm}
    \vskip 2em
  \end{center}
  \setcounter{footnote}{0}
}
\makeatother

\title{\LARGE \bf End-to-End Visual Autonomous Parking via Control-Aided Attention} 
\usepackage{amsmath,amsfonts}
\usepackage{array}
\usepackage{textcomp}
\usepackage{stfloats}
\usepackage{url}
\usepackage{verbatim}
\usepackage{graphicx}
\usepackage{cite}
\usepackage{graphics} 
\usepackage{epsfig} 
\usepackage{empheq}
\usepackage{times} 
\usepackage{amssymb}  
\usepackage{bm}
\usepackage{color}
\usepackage{soul}
\usepackage{bbding} 
\usepackage{diagbox}
\usepackage[table]{xcolor}
\usepackage{microtype} 
\usepackage{siunitx}

\usepackage[linesnumbered,ruled]{algorithm2e}
\usepackage{ulem} 
\usepackage{hyperref}
\hypersetup{
colorlinks=true,
linkcolor=blue,
filecolor=blue,      
urlcolor=blue,
citecolor=cyan,
}

\usepackage{float}
\usepackage{booktabs}
\usepackage{multirow}
\usepackage{mathrsfs}
\usepackage[utf8]{inputenc}
\usepackage{pifont}
\usepackage{threeparttable}
\usepackage{caption}
\usepackage{setspace}
\usepackage{balance}
\newcounter{RNum}

\usepackage{xcolor}
\usepackage{subcaption}
\usepackage{comment}
\usepackage{wrapfig}
\usepackage{pifont}
\usepackage{dsfont}

\definecolor{feng}{RGB}{0,128,255} 

\begin{document}


\author{
Chao Chen$^1$, Shunyu Yao$^1$, Yuanwu He$^1$, Feng Tao$^2$, Ruojing Song$^1$, 
Yuliang Guo\textsuperscript{1,\ding{105}}, Xinyu Huang$^2$, Chenxu Wu$^2$, 
Chen Feng\textsuperscript{1,\ding{41}}, and Liu Ren$^2$%
\thanks{\ding{41} Corresponding author.}
\thanks{\ding{105} Project lead.}
\thanks{$^{1}$Chao Chen, Shunyu Yao, Yuanwu He, Ruojing Song, and Chen Feng are with New York University.}%
\thanks{$^{2}$Feng Tao, Yuliang Guo, Xinyu Huang, Chenxu Wu, and Liu Ren are with Bosch Research.}
}


\maketitle

\thispagestyle{empty}
\pagestyle{empty}

\providecommand{\titlevariable}{CAA-Policy}

\begin{abstract}
Precise parking requires an end-to-end system where perception adaptively provides policy-relevant details—especially in critical areas where fine control decisions are essential.
End-to-end learning offers a unified framework by directly mapping sensor inputs to control actions, but existing approaches lack effective synergy between perception and control. 
Instead, we propose CAA-Policy, an end-to-end imitation learning system that allows control signal to guide the learning of visual attention via a novel Control-Aided Attention (CAA) mechanism. 
We train such an attention module in a self-supervised manner, using backpropagated gradients from the control outputs instead of from the training loss. This strategy encourages attention to focus on visual features that induce high variance in action outputs, rather than merely minimizing the training loss—a shift we demonstrate leads to a more robust and generalizable policy.
To further strengthen the framework, \titlevariable~incorporates short-horizon waypoint prediction as an auxiliary task to improve temporal consistency of control outputs, a learnable motion prediction module to robustly track target slots over time, and a modified target tokenization scheme for more effective feature fusion.
Extensive experiments in the CARLA simulator show that \titlevariable~consistently surpasses both the end-to-end learning baseline and the modular BEV segmentation + hybrid A*  pipeline, achieving superior accuracy, robustness, and interpretability. Code and Collected Training datasets will be released.
Code is released at \url{https://github.com/ai4ce/CAAPolicy}.
\end{abstract}


\section{Introduction}\label{sec:intro}

Autonomous driving has traditionally been approached with modular pipelines that decompose the task into perception, prediction, and planning/control~\cite{grigorescu2020survey,chen2024end,chib2023recent}. While such designs provide interpretability and modularity~\cite{chen2015deepdriving,maddern20171,kong2015kinematic}, they are prone to error accumulation across stages~\cite{wu2022trajectory}, redundant computation~\cite{kong2015kinematic,zhao2012design}, and reduced robustness under perturbations~\cite{wu2023adversarial}, \\Fig.~1\ref{fig:teasing}(a). End-to-end (E2E) learning offers a unified alternative by mapping sensory inputs directly to control outputs through a single neural network~\cite{hu2023planning,chitta2022transfuser,shao2023safety}, Fig.~1\ref{fig:teasing}(b). 

Existing end-to-end (E2E) autonomous parking methods~\cite{yang2024e2e} typically predict control commands directly from multi-view images. While these approaches unify perception and control, their synergy remains weak: perception modules are trained solely through gradients backpropagated from task losses, and thus often fail to focus on control-relevant cues such as parking slot boundaries, lane markings, and obstacle gaps. This limitation is particularly pronounced in attention mechanisms. As shown in Fig.~\ref{fig:caa_attn}(a–b), a baseline without control guidance scatters attention across irrelevant regions and exhibits temporal inconsistency. Moreover, even when incorporating an automatic learned attention module such as CBAM\cite{woo2018cbam}, the model still fails to consistently emphasize near-target areas. Without stronger coupling between perception and control, attention tends to drift toward spurious features, resulting in misaligned perception and unreliable policy outputs.

To address these limitations, we propose \titlevariable, an end-to-end autonomous driving framework that tightly couples perception and control (Fig. 1\ref{fig:teasing}(c)). At its core, it introduces a self-supervised Control-Aided Attention (CAA) mechanism that learns attention from control-output gradients rather than the loss gradients, enabling the attention module to consistently focus on control-sensitive regions near the target slot (Fig.~\ref{fig:caa_attn}(c)). Driving goals are further encoded as structured, interpretable tokens for goal-aware control, and short-horizon waypoint prediction is added as an auxiliary task to improve temporal consistency. Together, these components yield more accurate and robust autonomous parking.
\setcounter{figure}{1}
\begin{figure}[t]
    \centering
    \includegraphics[width=0.45\textwidth]{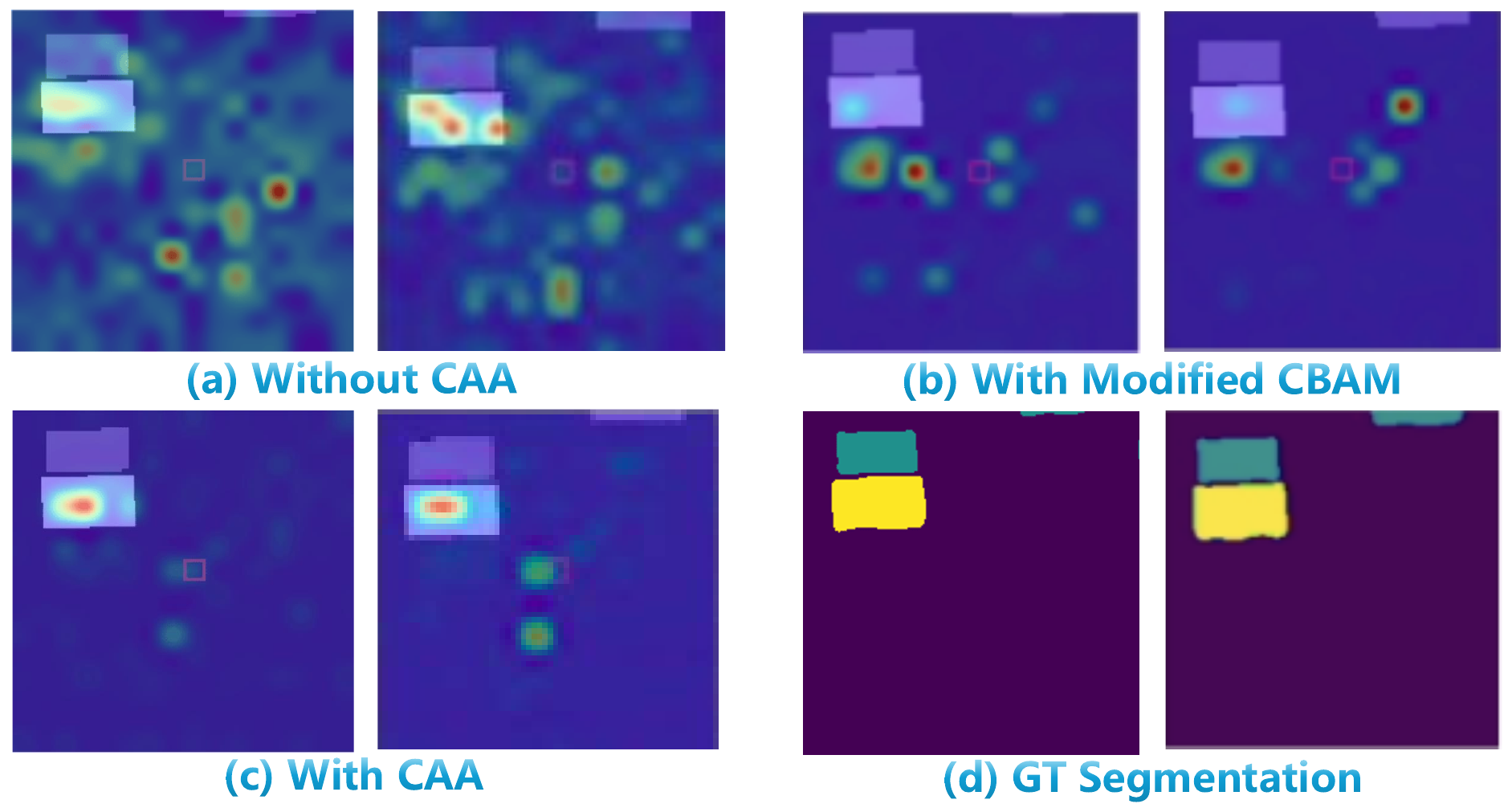}
    \vspace{-2mm}\caption{Decoder self-attention maps for target tracking at two consecutive frames ($t$ and $t{+}1$). 
(a) Without CAA, self-attention is scattered over irrelevant areas; 
(b) With modified CBAM~\cite{woo2018cbam}, an automatically learned attention module from task supervision (no auxiliary loss), attention is partially focused; 
(c) With CAA, self-attention is concentrated on the target and its vicinity; 
(d) Ground-truth segmentation (yellow: target, blue: obstacles) for reference.}
    \label{fig:caa_attn}

    \vspace{-6mm}
\end{figure}

\noindent
In summary, this paper makes the following contributions:
\begin{itemize}
\item An end-to-end visual autonomous parking framework, \textit{\titlevariable}, that features a novel \textit{Control-Aided Attention (CAA)} mechanism, which learns attention from control gradients on perception features, consistently focusing on control-sensitive areas.

\item We further improve performance by (1) introducing a learnable \textit{motion prediction} to reliably track the target spot, (2) improving \textit{target tokenization} to effectively encode the target input, and (3) adding a \textit{waypoint prediction} auxiliary task to enhance planning stability.




\item Through extensive experiments in CARLA, we demonstrate that \textit{\titlevariable}~ significantly outperforms both modular hybrid A* and previous end-to-end baselines in trajectory accuracy, success rate, and interpretability.
\end{itemize}

\section{Related Works }\label{sec:related}

\subsection{Autonomous parking system}

Prior work on autonomous parking ranges from traditional modular pipelines to modern End-to-End (E2E) methods. Early systems~\cite{chirca2015autonomous,kim2025robust,zhang2025enhanced} typically combined occupancy mapping, hand-crafted planners (e.g., hybrid A*~\cite{dolgov2010path}, RRT~\cite{lavalle1998rapidly}), and rule-based controllers, offering interpretability but limited adaptability. Recent advances largely adopt either modular learning approaches or E2E learning paradigms~\cite{al2025reinforcement,yang2024e2e,li2024parkinge2e}.

\textbf{Modular approaches} follow the classic decomposition of perception, planning, and control, with each stage designed and tuned separately. Modular pipelines with hybrid A* planners~\cite{dolgov2008practical, kurzer2016path} face fundamental efficiency limitations that directly impact accuracy under practical trade-offs. Owing to the breadth-first-search nature of hybrid A*, planning becomes increasingly costly when the target is far, as the frontier space grows with each step. To mitigate the high computational burden, systems often reduce the frequency of re-planning with new perception inputs. However, this leads to blind execution based on outdated information, where occlusions near the target are frequently ignored, resulting in frequent failures. This shortcoming has motivated the subsequent preference for learning-based policies and end-to-end approaches.

\textbf{End-to-end approaches} provide a unified alternative to modular pipelines by directly mapping sensory inputs to control outputs, making them particularly suitable for structured tasks like autonomous parking. Within this paradigm, one line of work formulates parking as a trajectory prediction task, where CNN–LSTM or Transformer architectures are supervised with trajectory waypoints rather than direct control commands~\cite{shen2020parkpredict,shen2022parkpredict+,li2024parkinge2e}. Another line of work adopts imitation learning to predict low-level control actions directly. For example, E2EParking~\cite{yang2024e2e} maps multi-view images to control outputs, but encodes goal semantics only implicitly within segmentation labels, which limits generalization. To enhance goal awareness, later methods~\cite{li2024parkinge2e,fu2025parkformer} explicitly represent parking slots—either by introducing slot-map queries fused with BEV features or by applying cross-attention between goal points and perception features.
Despite these variations, previous end-to-end methods lack explicit attention mechanisms to ensure perception focuses on critical regions for planning and control. We address this with Control-Aided Attention (CAA), which is trained to highlight control-sensitive areas.

\subsection{Deep Learning-based Motion Prediction} 

In our end-to-end parking setup, motion prediction is required to reliably track the ego vehicle’s state across multiple steps, ensuring that the planner operates on temporally consistent information. Prior end-to-end approaches such as E2EParking~\cite{yang2024e2e} employed segmentation-based tracking, but this design has been shown to fail in maintaining consistent predictions over multiple steps. Most traditional prediction methods are generally suitable for simple short-term tasks ~\cite{prevost2007extended, yuan2021agentformer,huang2022survey}, relying solely on the current state without modeling temporal dependencies. Recent temporal hybrid approaches combine multiple historical frames with physics-based constraints, employing architectures such as physics-informed RNN/LSTM~\cite{zyner2018recurrent,altche2017lstm,park2018sequence}, temporal CNN~\cite{phan2020covernet,gilles2021home}, CNN–RNN hybrids~\cite{deo2018convolutional,chandra2019traphic,xie2020motion}, and attention mechanisms~\cite{kim2020multi,messaoud2020attention,giuliari2021transformer} to enhance long-horizon accuracy while maintaining physically feasible trajectories. Since our scenario is a static simulated environment, we adopt a lightweight CNN combined with a single-layer LSTM model. This design ensures model efficiency while leveraging temporal information to improve prediction generalization across different policies.

\vspace{-1mm}
\subsection{Trajectory as Planning Output}
Trajectory-based methods are common in autonomous parking, as they explicitly generate a feasible path from the vehicle’s pose to a target slot. 
Recent works integrate learning modules to enhance robustness and comfort, with most models \cite{du2025transparking,shen2022parkpredict+,li2024parkinge2e} directly predicting trajectories from observations for controller execution. 
Alternatively, TCP \cite{wu2022trajectory} uses predicted trajectories as intermediate guidance for control. 
Building on this idea, \titlevariable~employs trajectory prediction as an auxiliary supervisory signal, benefiting from its structural guidance without requiring online replanning.

\providecommand{\mo}{\mathbf{o}}
\providecommand{\mq}{\mathbf{q}}
\newcommand{\lc}{\left ( }
\newcommand{\rc}{\right ) }

\providecommand{\positiveset}{\mathcal{P}_{{\bf{\mq}}_i}}
\providecommand{\postemporal}{\mathcal{P}^t_{{\bf{\mq}}_i}}
\providecommand{\negtemporal}{\mathcal{N}^t_{{\bf{\mq}}_i}}
\providecommand{\negset}{\mathcal{N}_{{\bf{\mq}}_i}}
\providecommand{\potpos}{\mathcal{\tilde{P}}_{{\bf{\mq}}_i}}
\providecommand{\verpos}{\hat{\mathcal{P}}_{{\bf{\mq}}_i}}
\providecommand{\feature}{f_{\theta}}

\section{Methodology}\label{sec:method}

\begin{figure*}[t]
\vspace{-2mm}
    \centering
    \includegraphics[width=0.94\textwidth, trim=0 0 3 0, clip]{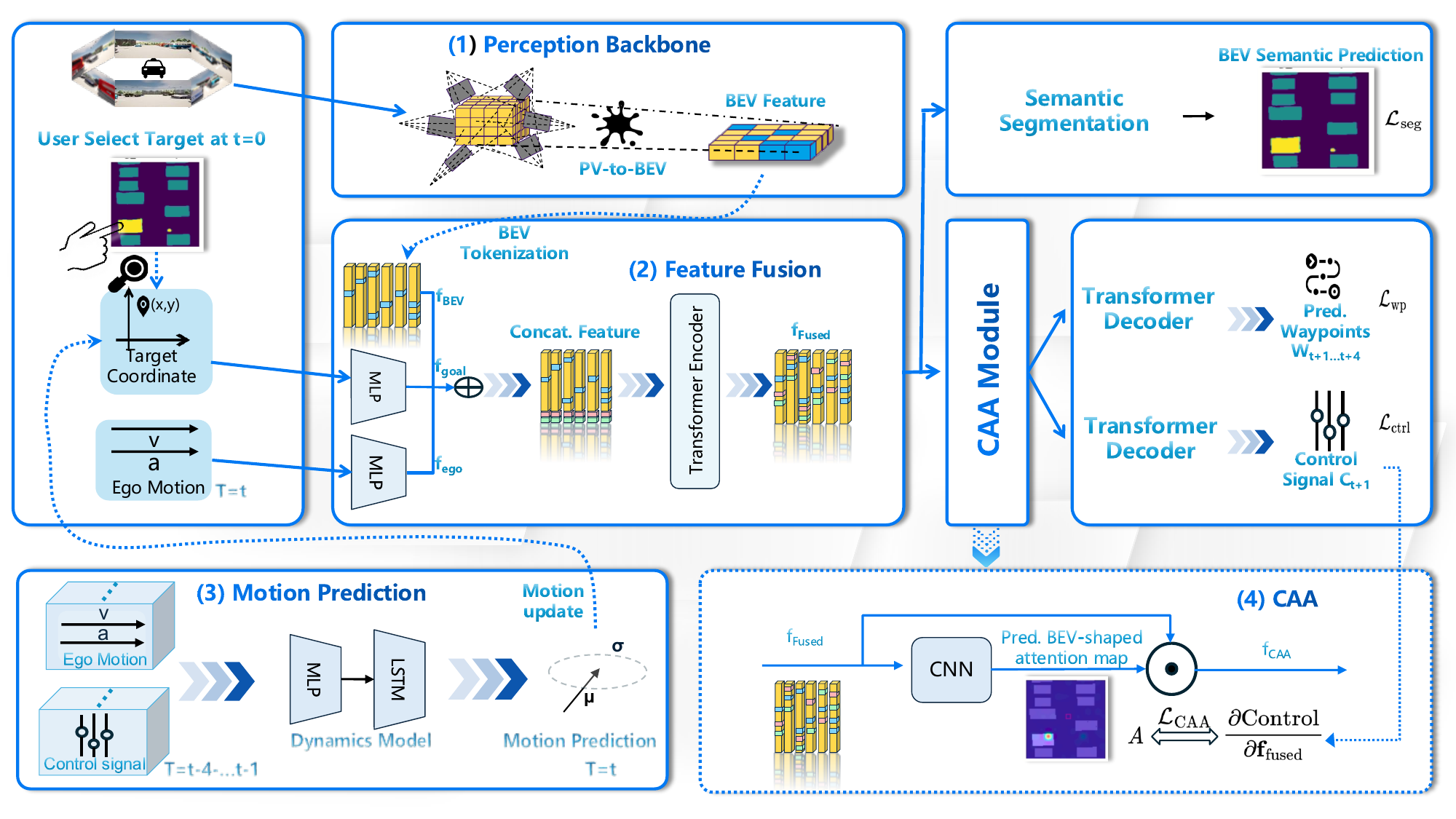} \quad
    \caption{\titlevariable~consists of five main components: 
\textbf{(1) Perception Backbone}, 
\textbf{(2) Feature Fusion},  
\textbf{(3) Learnable motion prediction module}, \textbf{(4) Control-Aided Attention (CAA) module} and 
\textbf{(5) Control and Waypoint Prediction}. 
The network also incorporates auxiliary heads for depth estimation and semantic segmentation, following the design of E2EParking~\cite{yang2024e2e}, to improve feature representation.
Notably, the Learnable motion prediction module is only used during inference, leveraging historical control and vehicle state information to reason about ego-vehicle dynamics and target position. }

    \label{fig:workflow}
\vspace{-5mm}
\end{figure*}

\subsection{Problem Statement}

The goal of \titlevariable~is to control an autonomous vehicle to park safely and accurately into a designated slot using end-to-end learning. At each time step $t$, the vehicle receives the following \textbf{inputs} \textit{during inference}:

\begin{itemize}
    \item Multi-view images $I_t \in \mathbb{R}^{6 \times H \times W \times 3}$ captured from six surround cameras covering the vehicle: front, front-left, front-right, rear, rear-left, and rear-right views.

    \item Current vehicle state $\mathbf{s}_t$, including scalar speed $v_t$, planar acceleration $\mathbf{a}_t$, and heading (yaw) $\psi_t$.
    \item Target parking goal $\mathbf{g}_t = (x_t^\text{target}, y_t^\text{target}, \psi_t^\text{target})$, in the ego frame at time $t$.
    \item Inputs to the motion prediction module: historical motion from four consecutive frames, control commands $\mathbf{c}_{t-4:t-1}$, and vehicle states $\mathbf{s}_{t-4:t-1}$.

\end{itemize}

The system \textbf{outputs} position, short-horizon control, and perception predictions:
\begin{itemize}
    \item Current-frame position $(x_t, y_t)$, predicted by the motion prediction module and used to update the ego-frame target coordinates for this step.

    \item Short-horizon control commands for the next 4 steps, 
$\mathbf{c}_{t+1:t+4} = (\text{steering}, \text{throttle}, \text{brake})_{t+1:t+4}$.

    \item Short-horizon 2D ego-centric waypoints for the next 4 time steps, used to ensure smooth and consistent future trajectory: $\mathbf{w}_{t+1:t+4} = [(x_{t+1}, y_{t+1}, \psi_{t+1}), \dots, (x_{t+4}, y_{t+4}, \psi_{t+4})]$.
    \item Auxiliary outputs supporting perception tasks, including depth estimation and BEV semantics.

\end{itemize}

The objective is to train a supervised policy $\pi$ from expert trajectories 
$\mathcal{D}=\{\tau^n\}_{n=1}^N$, 
where $N$ is the number of expert trajectories. Each trajectory 
$\tau^n=\{(J^n_t, K^n_t)\}_{t=1}^{T_n}$ contains $T_n$ time steps:
\begin{itemize}
    \item $J^n_t$: multi-view images, vehicle state, target goal, and historical motion.
    \item $K^n_t$: expert labels for all supervised tasks (control, next pose, waypoints, depth, BEV semantics).
\end{itemize}

The policy $\pi$ is optimized to minimize
\vspace{-1mm}
\[
\pi^* = \arg \min_\pi \frac{1}{N} \sum_{n=1}^N \frac{1}{T_n} \sum_{t=1}^{T_n} \mathcal{L}\left(K_t^n, \pi\left(J_t^n\right)\right),
\]
where $\mathcal{L}$ is a general supervised loss capturing all tasks.


\subsection{Network Architecture}
\titlevariable~consists of five main stages: (1) Perception Backbone, (2) Feature Fusion, (3) Learnable motion prediction module, (4) Control-Aided Attention(CAA) module,
and (5) Control and Waypoint Prediction, as illustrated in Fig.~\ref{fig:workflow}. Auxiliary heads for depth and semantic segmentation are also included, following the design of E2EParking~\cite{yang2024e2e}. Notions are explicitly shown in Tab.~\ref{tab:notation}

\begin{table}[h]
\centering

\captionsetup{font=sc}
\captionsetup{font={scriptsize, sc, stretch=1.3}, justification=centering, labelsep=newline}

\caption{Summary of major notations for \titlevariable, grouped by module. 
The Symbol column has a light gray background for easy distinction, while module headers have a very light blue background. 
Notation conventions: continuous quantities are denoted by $\tilde{\cdot}$ (e.g., $\tilde{\mathbf{w}}, \tilde{\mathbf{c}}$); 
ground-truth values are denoted by $^\ast$ (e.g., $\mathbf{w}^\ast, \mathbf{c}^\ast$); 
predicted distributions are denoted by $\hat{\cdot}$ (e.g., $\hat{\mathbf{w}}, \hat{\mathbf{c}}$).}
\resizebox{0.48\textwidth}{!}{ 
\begin{tabular}{>{\columncolor{gray!10}}l l} 
\toprule
\rowcolor{blue!5} 
\multicolumn{2}{c}{\textls[200]{\textit{Perception}}} \\
$I_t$ & Multi-view images at time $t$ \\
$\mathbf{f}_{\text{BEV}}$ & Tokenized BEV feature from images \\
$\mathbf{f}_{\text{ego}}$ & Tokenized ego vehicle state vector \\
$\mathbf{f}_{\text{goal}}$ & Tokenized parking goal vector \\
$\mathbf{f}_{\text{fused}}$ & Concatenated Tokenized feature vector \\
\midrule
\rowcolor{blue!5} 
\multicolumn{2}{c}{\textls[200]{\textit{Planning \& Motion Prediction}}} \\
$\mathbf{c}_{t+1:t+4}$ & Short-horizon control command vectors (steering, throttle, brake) \\
$\mathbf{w}_{t+1:t+4}$ & Short-horizon future 2D waypoint vectors \\
$(x_{t},y_{t})$ & Predicted current vehicle position (2D scalar coordinates) \\
$\mathbf{h}_t$ & Dynamics LSTM hidden state vector \\
$\boldsymbol{\mu}_t, \boldsymbol{\sigma}^2_t$ & Predicted 2D displacement mean \& variance vectors \\
\midrule
\rowcolor{blue!5} 
\multicolumn{2}{c}{\textls[200]{\textit{Attention \& Losses}}} \\
$\mathbf{A}$ & Predicted BEV attention map (CAA) \\
$\nabla_{\mathbf{f}}^\ast$ & Ground-truth control gradient w.r.t. fused feature\\

$\mathbf{f}_{\text{CAA}}$ & CAA-refined BEV feature vector \\
$\mathcal{L}_\text{c}, \mathcal{L}_\text{w}, \dots$ & Task-specific losses \\
\bottomrule
\end{tabular}}

\label{tab:notation}
\end{table}

\subsubsection{Perception backbone}\label{sec:perception}
We adopt a ResNet-18~\cite{he2016deep} encoder shared across six surround-view RGB images. The resulting features are fused into a unified bird’s-eye view (BEV) representation using the Lift-Splat-Shoot (LSS) framework~\cite{philion2020lift}. To regularize the backbone and ensure the BEV features capture structural cues beyond control supervision, we employ auxiliary heads for semantic segmentation and depth estimation, ensuring the BEV features encode both geometric and semantic cues useful for downstream modules.

\subsubsection{Feature Fusion with Target Tokenization Module (TTM)}
In addition to the BEV feature extracted from multi-view images, vehicle state and parking goal information are required for downstream planning.
Compared to previous end-to-end parking approaches~\cite{yang2024e2e}, which embed the target slot implicitly into the BEV feature before tokenization, we explicitly tokenize all input information. The BEV features and vehicle states are encoded via CNN and MLP modules into compact representations, and the ego-target coordinates are explicitly tokenized. These representations are then concatenated to form the final fused feature:
\vspace{-1mm}
\begin{equation}
\begin{aligned}
\mathbf{f}_{\text{BEV}}, \mathbf{f}_{\text{ego}}, \mathbf{f}_{\text{goal}} &= 
\mathrm{CNN}(\mathbf{I}_{1:N}), \mathrm{MLP}(\mathbf{s}), \mathrm{MLP}(\mathbf{g}),\\
\mathbf{f}_{\text{fused}} &= \mathrm{Concat}(\mathbf{f}_{\text{BEV}}, \mathbf{f}_{\text{ego}}, \mathbf{f}_{\text{goal}})
\end{aligned}
\end{equation}
\vspace{-1mm}

where $\mathbf{I}_{1:N}$ denotes the set of $N$ multi-view images, 
$\mathbf{s}$ represents the vehicle state, 
$\mathbf{g}$ denotes the target parking slot coordinates in the ego-vehicle frame, 
$\mathbf{f}_{\text{BEV}}$, $\mathbf{f}_{\text{ego}}$, and $\mathbf{f}_{\text{goal}}$ are their corresponding tokenized features, 
and $\mathbf{f}_{\text{fused}}$ is the concatenated fused representation.

The encode the target effectively, we explicitly convert the target parking slot coordinates into a token via $\mathbf{f}_{\text{goal}} = \text{MLP}(\mathbf{g})$. This explicit encoding preserves geometric information about the goal more effectively, improving trajectory prediction and control performance. The effect of this design is further analyzed in Sec.~\ref{sec:ablation}.



\subsubsection{Learnable Motion Prediction}

We leverage the previous four steps of control commands $\mathbf{c}_{t-4:t-1}$ and vehicle states $\mathbf{s}_{t-4:t-1}$ to estimate the 2D displacement relative to the last frame. Frame-wise features are encoded by a CNN and aggregated by an LSTM into the temporal latent state $h_t = \mathrm{LSTM}(\{\mathrm{CNN}([\mathbf{c}_k, \mathbf{s}_k])\}_{k=t-4}^{t-1})$. Two MLP heads predict the displacement mean and variance vectors $\boldsymbol{\mu}_t$ and $\boldsymbol{\sigma}_t^2$, and the vehicle position is updated as $(x_t, y_t) = (x_{t-1}, y_{t-1}) + \boldsymbol{\mu}_t$, which refreshes the ego-frame target coordinates. Here, $\boldsymbol{\mu}_t$ and $\boldsymbol{\sigma}_t^2$ represent per-axis displacement mean and variance.

During training, the predicted displacement $\boldsymbol{\mu}_t$ is supervised by the ground-truth displacement $\Delta \mathbf{p}_t^\ast = (x_t^\ast - x_{t-1}^\ast, \; y_t^\ast - y_{t-1}^\ast)$ using the NLL loss: $\mathcal{L}_{\text{dyn}} = - \sum_{i \in \{x, y\}} \log \mathcal{N}(\Delta p_{t,i}^\ast \mid \mu_{t,i}, \sigma_{t,i}^2)$, where each axis is treated independently.

\textbf{Remark.} This module is only used during inference, and itself is trained separately; during the end-to-end training of the major perception and planning system, ground-truth positions are directly used.

\subsubsection{Control and Waypoint Prediction}

Both control and waypoint prediction modules take the CAA-refined features 
\(\mathbf{f}_{\text{CAA}} \in \mathbb{R}^{C_b \times X_b \times Y_b}\) as input. 
The control prediction head follows the structure of E2EParking~\cite{yang2024e2e}, 
producing discrete control tokens autoregressively. Continuous control signals (steering, throttle, and brake) are discretized using the same scheme as in E2EParking, with identical bin boundaries, minimum, and maximum values for each control dimension. 

The waypoint prediction head explicitly models short-horizon 2D vehicle positions. 
Using only the current \(\mathbf{f}_{\text{CAA}}\), it outputs future waypoints for the next few steps:
$
\mathbf{w}_{t+1:t+4} = \mathrm{WaypointDecoder}(\mathbf{f}_{\text{CAA}}),
$
where \(\mathbf{w}_{t+1:t+4}\) are the predicted waypoints in ego-centric coordinates. 
The XY positions and heading angles are discretized using boundaries determined by statistical analysis of all training and testing trajectories: 
$\Delta x$ and $\Delta y$ are bounded within [-6, 6] meters, and the heading change $\Delta \psi$ is bounded within [-40, 40] degrees, with each dimension uniformly divided into 200 bins. 
These ego-centric waypoints are then transformed into global coordinates using the current vehicle pose at time $t$.

This design allows both control and waypoint heads to leverage task-specific, 
CAA-refined perception features. The additional waypoint supervision provides explicit guidance on vehicle geometry and motion, improving trajectory stability and control accuracy.

\subsubsection{Control-Aided Attention (CAA)}\label{sec:CAA}

The Control-Aided Attention (CAA) encourages the attention module to focus on control-sensitive areas, allowing the perception module to provide more reliable information for planning and control. For each BEV location, CAA predicts the expected control gradient in a self-supervised manner. 

Given fused tokens $\mathbf{f}_{\text{fused}}$, we first encode them with a CNN to obtain a BEV feature map
\[
\mathbf{b}_{\text{BEV}} = \mathrm{CNN}(\mathbf{f}_{\text{fused}}) \in \mathbb{R}^{C_b \times X_b \times Y_b},
\]
where $C_b$ is the number of channels and $X_b, Y_b$ are the spatial dimensions of the BEV grid. CAA then outputs an attention map 
$
A \in \mathbb{R}^{X_b \times Y_b},
$
which refines the BEV features by spatially weighting each location (broadcasted across channels):
\begin{equation}
\begin{aligned}
\mathbf{f}_{\text{CAA}}[:,i,j] &= \mathbf{b}_{\text{BEV}}[:,i,j] \cdot A[i,j],\\
 &i=1,\dots,X_b,\; j=1,\dots,Y_b.
\end{aligned}
\end{equation}

Gradients from the predicted control signals guide the attention, highlighting regions that most influence future actions and improving the robustness and generalization of downstream planning and control.


\vspace{-1mm}
\subsection{Loss Function}\label{sec:loss}

To achieve robust autonomous driving in complex urban scenes, \titlevariable~is trained under a unified multi-task framework that jointly optimizes control, waypoint prediction, perception, and self-supervised attention alignment.
Complementary objectives regularize each other: control and waypoint prediction benefit from perception (segmentation and depth), while the Control-Aided Attention (CAA) loss aligns perception features with decision-making signals.
The overall training objective integrates these components into a single loss:
\begin{equation}
\mathcal{L}_\text{total} = \mathcal{L}_\text{c} 
+ \mathcal{L}_\text{w} 
+  \mathcal{L}_\text{s} 
+  \mathcal{L}_\text{d} 
+  \mathcal{L}_\text{C}.
\end{equation}

\paragraph{Control Loss} 
at each time step $t$, the ground-truth continuous control command 
$
\tilde{c}_t^{^\ast} = (\text{steering}_t, \text{throttle}_t, \text{brake}_t)$ is uniformly quantized into a discrete token index $c_t^{\ast}$. 
Let $\hat{c}_t$ be the predicted token distribution over the discrete control vocabulary. 
The control loss is then defined as:
\vspace{-1mm}
\begin{equation}
\mathcal{L}_\text{c} = \sum_{k=1}^{4} \text{CrossEntropy}(\hat{c}_{t+k}, c_{t+k}^{\ast}),
\end{equation}
\vspace{-1mm}

where $c_{t+k}^{\ast}$ denotes the ground-truth discrete control token at the $k$-th future step and $\hat{c}_{t+k}$ is the corresponding predicted probability distribution.

\paragraph{Waypoint Loss} 
at each time step $t$, the short-horizon future continuous waypoints 
$
\tilde{\mathbf{w}}_{t+1:t+4}^{\ast} = 
[(x_{t+1}, y_{t+1}, \psi_{t+1}), \dots, (x_{t+4}, y_{t+4}, \psi_{t+4})]
$
are uniformly quantized into discrete token indices 
$\mathbf{w}_{t+1:t+4}^{\ast} = [w_{t+1}^{\ast}, \dots, w_{t+4}^{\ast}]$.  
Let $\hat{w}_{t+k}$ be the predicted token distribution for the $k$-th future waypoint.  
The waypoint loss is then defined as:
\vspace{-1mm}
\begin{equation}
\mathcal{L}_\text{w} = \sum_{k=1}^{4} 
\text{CrossEntropy}(\hat{w}_{t+k}, w_{t+k}^{\ast}),
\end{equation}
where $w_{t+k}^{\ast}$ is the ground-truth discretized waypoint, and $\hat{w}_{t+k}$ the predicted probability distribution.

\paragraph{Perception Losses} 
Following E2EParking~\cite{yang2024e2e}, we supervise BEV semantic segmentation with cross-entropy 
$\mathcal{L}_\text{s} = \text{CE}(\hat{M}, M)$ 
and depth prediction with binary cross-entropy 
$\mathcal{L}_\text{d} = \tfrac{1}{N_\text{f}} \sum_{i \in \mathcal{FG}} \text{BCE}(\hat{D}_i, D_i),$ where $\hat{M}$ and $\hat{D}$ are predicted BEV semantic and depth maps, 
$M$ and $D$ are ground-truth maps, and $N_{\mathcal{FG}}$ is the number of foreground pixels $\mathcal{FG}$ with valid depth.


\paragraph{Control-Aided Attention (CAA) Loss} 
CAA module predicts a BEV-shaped attention map $A \in \mathbb{R}^{H \times W}$ that approximates the control sensitivity of the corresponding BEV location. Inspired by Grad-CAM~\cite{selvaraju2017gradcam}, during training, the ground-truth control gradient is obtained via automatic differentiation:
$
\nabla_f^\ast = \partial \text{Control} / {\partial \mathbf{f}_{\text{fused}}}.
$

The prediction is then supervised with an $L_1$ loss:
\vspace{-2mm}

\[
\mathcal{L}_{\text{C}} = \left\| A - \nabla_f^\ast \right\|_1.
\]
\vspace{-2mm}

This self-supervised loss encourages the perception module to focus on regions that most strongly affect downstream control predictions.

\section{Experiments}\label{sec:experiment}

Experiments are conducted in the CARLA 0.9.11 simulator (Town04-Opt), a parking lot scenario with 64 slots, following the extended environment of E2EParking~\cite{yang2024e2e}.
Training and validation data are collected from successful parking trajectories generated by traditional modular pipelines with ground-truth BEV segmentation inputs, starting from random positions towards specified targets. For testing, we evaluate parking in unseen scenarios around novel parking slots. Model robustness is assessed using TSR, TFR, NTSR, CR and TR metrics, as reported in Tab.~\ref{tab:evaluate}.

\begin{table}[t]
\centering
\captionsetup{font=sc}
\captionsetup{font={scriptsize, sc, stretch=1.3}, justification=centering, labelsep=newline}

\caption{Key evaluation metrics for autonomous parking.}
\resizebox{0.94\columnwidth}{!}{%
\begin{tabular}{>{\columncolor{gray!15}}p{0.18\columnwidth} p{0.75\columnwidth}}
\toprule
\textbf{Metric} & \textbf{Description} \\
\midrule
TSR & \textit{Target Success Rate.} Probability that the ego vehicle successfully parks into the target slot. Success: lateral distance $<0.6$ m, longitudinal distance $<1$ m, orientation error $<10^\circ$. \\[0.8ex]
TFR & Target Failure Rate. Probability that the ego vehicle parks into the target slot but with unacceptable error. \\[0.8ex]
NTSR & \textit{Non-Target Success Rate.} Probability that the ego vehicle parks into a non-target slot. \\[0.8ex]
CR & \textit{Collision Rate.} Probability of collision with other obstacles. \\[0.8ex]
TR & \textit{Timeout Rate.} Probability that the ego vehicle fails to park within 40 second. \\
\bottomrule
\end{tabular}%
}
\label{tab:evaluate}
\vspace{-4mm}
\end{table}

\subsection{Implementation details}
Most of our implementation follows E2EParking~\cite{yang2024e2e}, including data collection, BEV feature size and the corresponding spatial resolution, backbone architecture, depth image parameters, encoder–decoder attention setup, and the optimization framework.

Compared to E2EParking, our setup mainly differs in dataset scale and training configuration. 
The training set contains $820$ trajectories ($187{,}379$ frames) and the validation set includes $175$ tasks ($40{,}225$ frames), both pre-collected for training the model. 
Each trajectory averages about $200$ frames, collected at $30$ Hz, while the model processes inputs at $0.1$ s intervals (every three steps). 
We train for $20$ epochs ($\approx$48 hours on 4 NVIDIA RTX 8000 GPUs with batch size $32$). 
Additionally, we use $6$ surrounding cameras (instead of $4$) to better cover blind spots. 
Testing is conducted in a separate environment by running complete parking episodes.

For expert demonstrations, rather than using human driving as in~\cite{yang2024e2e}, 
we employ trajectories generated by a modular Hybrid A* planner, with control signals obtained from a PID-controlled bicycle model. 
Only successful and smooth trajectories are manually selected for training.


\vspace{-1mm}
\subsection{Baselines}
\vspace{-1mm}

We compare \titlevariable~against two baselines: E2E baseline~\cite{yang2024e2e} and Modular Hybrid A* baseline.
\begin{table*}[t]
\centering
\small
\renewcommand{\arraystretch}{1.5}
\captionsetup{font=sc}
\captionsetup{font={scriptsize, sc, stretch=1.3}, justification=centering, labelsep=newline}

\caption{Comparison of parking performance across E2E baselines, ablations, the modular baseline, and the full model (Ours). 
All results are reported as $\mu \pm \sigma$ (mean $\pm$ standard deviation) in percentage over evaluation episodes. 
Metric definitions are provided in Table~\ref{tab:evaluate}; minor errors omitted.}

\label{tab:ablation}
\resizebox{\textwidth}{!}{%
\begin{tabular}{lcccccccccc}
\hline
\textbf{Methods} & \textbf{MotionPred} & \textbf{TTM} & \textbf{WayPts} & \textbf{CAA} & \textbf{TSR$\uparrow$} & \textbf{TFR$\downarrow$} & \textbf{NTSR$\downarrow$} & \textbf{CR$\downarrow$} & \textbf{TR$\downarrow$} \\
\hline
Hybrid A*~\cite{dolgov2008practical, kurzer2016path} & & & & &  59.1 $\pm$ 3.8 &  5.5 $\pm$ 2.2   &  0.0 $\pm$ 0.0   &  12.2 $\pm$ 2.9   &  21.1 $\pm$ 3.7 \\
\hline
E2EParking (Checkpoint)~\cite{yang2024e2e} & & & & &  74.7 $\pm$ 5.0 &  1.0 $\pm$ 0.6 &  3.1 $\pm$ 1.3 &  12.0 $\pm$ 3.8 &  8.9 $\pm$ 3.3 \\

E2EParking (Reproduced)~\cite{yang2024e2e}  & & & & &  17.6 $\pm$ 6.3 & 2.9 $\pm$ 1.7 &  29.4 $\pm$ 2.9 &  35.3 $\pm$ 5.5 &  14.7 $\pm$ 3.4 \\

\hline
\multirow{7}{*}{\titlevariable~(Ablation)} 
&  \checkmark & & & &  26.8 $\pm$ 5.1 &  3.5 $\pm$ 1.7 &  19.9 $\pm$ 3.4 &  34.2 $\pm$ 5.2 &  10.4 $\pm$ 3.2 \\
& \checkmark & \checkmark & & &  74.0 $\pm$ 5.7 &  1.8 $\pm$ 1.6 &  0.3 $\pm$ 0.2 &  13.5 $\pm$ 3.9 &  9.9 $\pm$ 3.2 \\
& \checkmark & \checkmark &  & \checkmark &  81.1 $\pm$ 4.5 &  4.5 $\pm$ 1.9 &  0.2 $\pm$ 0.2 &  4.4 $\pm$ 2.0 &  8.7 $\pm$ 3.2 \\
& \checkmark & & \checkmark &  &  76.0 $\pm$ 6.0 &  0.8 $\pm$ 0.5 &  0.0 $\pm$ 0.0 &  9.1 $\pm$ 3.6 &  12.2 $\pm$ 3.8 \\
& \checkmark & & \checkmark &  \checkmark &  80.2 $\pm$ 4.9 &  2.8 $\pm$ 1.6 &  5.9 $\pm$ 1.9 &  7.4 $\pm$ 2.8 &  2.1 $\pm$ 1.2 \\
& \checkmark & \checkmark & \checkmark &  &  81.0 $\pm$ 5.4 &  2.3 $\pm$ 1.6 &  2.7 $\pm$ 0.8 &  6.2 $\pm$ 3.9 &  7.1 $\pm$ 3.4 \\
& \checkmark & \checkmark & \checkmark & CBAM~\cite{woo2018cbam} &  82.4 $\pm$ 4.3 &  2.1 $\pm$ 1.8 &  2.9 $\pm$ 0.9 &  7.1 $\pm$ 2.3 &  4.4 $\pm$ 1.9 \\
\hline
{\titlevariable~(Ours)} & \checkmark & \checkmark & \checkmark & \checkmark & \textbf{87.5 $\pm$ 3.7} & \textbf{2.3 $\pm$ 0.9} & \textbf{0.8 $\pm$ 0.6} & \textbf{3.5 $\pm$ 2.6} & \textbf{3.8 $\pm$ 2.2} \\
\hline
\end{tabular}%
}

\vspace{-4mm}
\end{table*}

\subsubsection{E2E baseline study}



We begin with the E2EParking baseline~\cite{yang2024e2e}, an end-to-end model that predicts control commands from raw multi-view images. Using the original checkpoint, it achieves a TSR of 74.7\%, supported by a substantially larger training set and longer training schedule than those available in our setting. In contrast, our reproduction with only 820 expert trajectories and half the training time attains a TSR of 17.6\% and a CR of 35.3\%. These results highlight the sensitivity of end-to-end methods to dataset scale: with limited data, The baseline struggles to generalize to unseen parking slots and fails to reliably track the target.


\subsubsection{Modular baseline study}
For comparison, we evaluate a modular autonomous driving pipeline using Hybrid A*~\cite{dolgov2008practical, kurzer2016path} for planning. Obstacles are derived from BEV segmentation, ego localization uses our motion prediction model, and a PID-controlled bicycle model handles vehicle control. This baseline achieves a TSR of 59.1\% with zero NTSR, but exhibits a TR of 21.1\% and CR of 12.2\%, indicating that modular pipelines, while stable, are limited in handling diverse or long-range parking scenarios.

\subsection{Failure case study}

\begin{figure*}[t]
    \centering
    \includegraphics[width=0.93\textwidth, trim=0 0 3 0, clip]{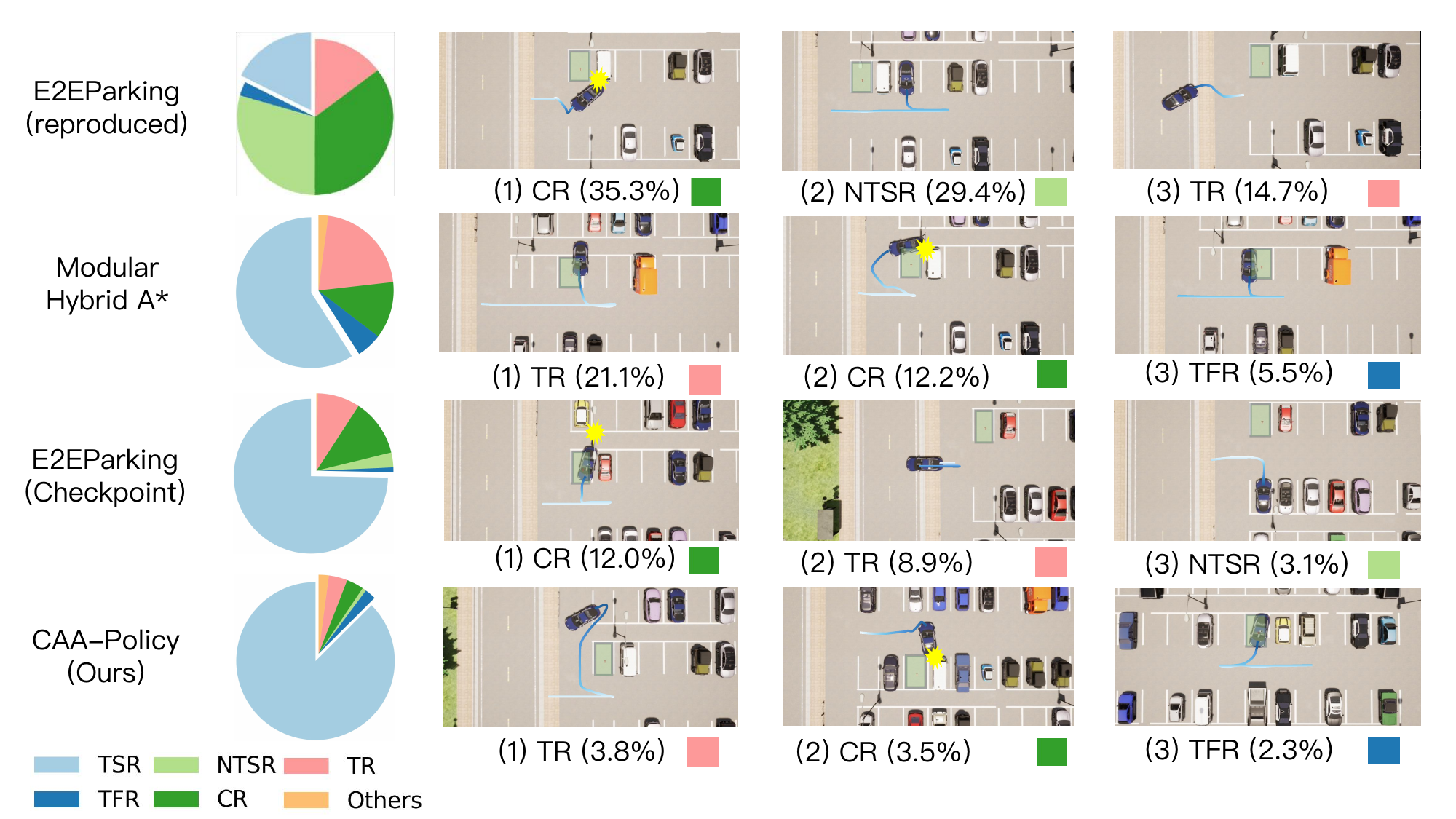}
    \caption{Failure case study across different baselines and our method. 
    Each row corresponds to a different experiment: 
    (1) reproduced E2EParking trained with limited data, 
    (2) Modular Hybrid A* baseline, 
    (3) original E2EParking checkpoint trained with large-scale data and longer schedule, 
    (4) \titlevariable~(Ours). 
    Each column shows: 
    (a) distribution of key metrics (TSR, TFR, NTSR, CR, TR) and minor errors (e.g., out-of-bound) as pie charts.
    (b-d) the top three representative failure modes observed in that experiment. 
    This layout allows direct comparison of overall performance and failure types across baselines and our method. Note that Trajectories are color-coded according to time sequence. For the Modular Hybrid A* baseline, red trajectories indicate the planned GT trajectory, which the controller struggles to follow on consecutive turns.}
    \label{fig:failure_study}
    \vspace{-3mm}
\end{figure*}


\textit{E2EParking}: In Fig.~\ref{fig:failure_study}, we analyze both the reproduced version trained with limited data (first row) and the original checkpoint trained on large-scale data (third row).
Our failure case analysis shows three dominant patterns: (i) collisions with nearby vehicles, (ii) misalignment to unintended slots, and (iii) timeouts where the vehicle fails to reach the target slot within the time limit.
These issues primarily arise from the reliance of E2EParking on segmentation-based estimation of the current target, which serves as a single point of failure. Small inaccuracies in target estimation can propagate through motion prediction, leading to larger trajectory deviations.
In the low-data regime, the same failure types persist but occur more frequently, underscoring the sensitivity of purely end-to-end policies to training scale.
These observations motivate our introduction of explicit modules (e.g., TTM and waypoints) to provide stronger target awareness and trajectory guidance.

\textit{Modular Hybrid A*}: As shown in the second row of Fig.~\ref{fig:failure_study}, despite the low NSTR suggesting that Hybrid A*~\cite{dolgov2008practical, kurzer2016path} can plan reliably toward the correct target, it frequently fails due to time run-outs and collisions. These failures primarily stem from three factors:
(1) Compounding perception errors: small inaccuracies in object detection or free-space estimation can misguide trajectory planning;
(2) Computation overhead: generating feasible paths to distant targets is computationally expensive, limiting the update frequency of perception inputs;
(3) Controller-induced tracking errors: executing trajectories with a separate controller (e.g., PID) introduces deviations that accumulate over long horizons. This issue is especially pronounced in scenarios with consecutive sharp turns, where even minor path errors can cause large deviations.
It is also worth noting that our evaluation directly integrates~\cite{dolgov2008practical, kurzer2016path} into the modular framework without extensive engineering of the number of planning and execution steps or manual tuning of search-tree expansion parameters—which could substantially improve real-world performance.
\textit{\titlevariable}: Finally, we analyze \titlevariable~(fourth row of Fig.~\ref{fig:failure_study}). Overall failure rates are significantly lower than other methods, with all error types controlled below 4\%, and the few remaining errors occurring near the target. This demonstrates that the learnable motion prediction module effectively follows the vehicle’s current state, and the Target Tokenization Module (TTM) accurately abstracts the parking goal, together ensuring precise target tracking. The Waypoint module further improves temporal consistency of predicted motions, while the Control-Aided Attention (CAA) mechanism emphasizes control-sensitive regions, enhancing attention to target cues and increasing parking success.


\subsection{Ablation Study}\label{sec:ablation}
We analyze the contribution of each key module in DL-based Motion Prediction (MotionPred), Target Tokenization Module (TTM), and Waypoint Prediction Head (WayPts), \titlevariable~- Control-Aided Attention (CAA)- and their combinations to quantify complementary benefits and overall performance gains shown in Table~\ref{tab:ablation}.


\subsubsection{DL-based Motion Prediction (MotionPred)} 
The failure analysis of both the original E2EParking and our reproduced version indicates that parking to an incorrect target is a critical issue. As shown, the introduced DL-based motion prediction module significantly reduce the NTSR and improve TSR. Given its essential role, this module is included as a default component in all remaining variants of our method.


\subsubsection{Target Tokenization Module (TTM)}
The TTM module raises TSR from 26.8\% (baseline with our motion prediction) to 74.0\%. By explicitly encoding the target parking slot, it clarifies the agent’s goal, stabilizing control and reducing ambiguous behaviors. It also lowers NTSR from 19.9\% to 0.3\%. Some failures remain—collision rate at 13.5\% and timeout rate at 9.9\%—showing TTM still benefits from complementary trajectory guidance modules.

\subsubsection{Waypoint module (WayPts)}  
When applying only the Waypoint module, the model achieves a TSR of 76.0\%, representing a substantial improvement over the reproduced baseline and matching the performance of the E2EParking checkpoint trained on large-scale data, while reducing the CR to 9.1\%. This indicates that explicit intermediate waypoints provide strong spatial guidance, constraining the agent’s trajectory and reducing unintended deviations.  However, the Timeout Rate remains relatively high (12.2\%).

\subsubsection{CAA module}
The CAA module refines perception features by emphasizing control-relevant regions. As shown in the ablation study, combining this module with either the TTM module, the Waypoint module, or both consistently improves performance. Since the other modules already elevate the overall system to a high baseline, achieving an additional gain of over 5\% in success rate is particularly valuable for meeting the demands of precision parking. We also observe that replacing CAA with CBAM~\cite{woo2018cbam} does not lead to notable improvements. This confirms the contribution of the control-sensitive supervison we introduced in CAA.

\subsubsection{Module Synergy}  
We investigate the synergy between TTM, Waypoints and CAA. Removing any single module while keeping the other two yields TSR values around 80–81\%, indicating complementary contributions. However, all partial combinations fall short of the full \titlevariable, which achieves 87.5\% TSR — roughly 6–7\% higher than any ablation — showing that jointly integrating TTM, Waypoints, and CAA maximizes performance and reduces collisions, timeouts, and non-target successes.

\vspace{-2mm}
\section{Limitations}

Despite \titlevariable's strong performance, limitations remain: evaluation excludes dynamic objects, limiting assessment in interactive scenarios; robustness to speed and target noise is partial, and environmental variations such as lighting and weather remain unaddressed; finally, as experiments are in simulation, real-world transfer requires further study.


\section{Conclusion}

In this work, we present \titlevariable, an end-to-end autonomous parking framework that integrates Control-Aided Attention (CAA) to couple perception with control demands, together with enhancements from a learnable motion prediction module, a Target Tokenization Module (TTM), and an auxiliary waypoint prediction task. Experiments in CARLA show that each module improves performance, while their combination yields substantial gains in trajectory stability, control accuracy, and success rate, demonstrating the effectiveness of incorporating modular designs into an end-to-end learnable system.


{\small
\bibliographystyle{IEEEtran}
\balance
\bibliography{egbib}
}

\clearpage
\renewcommand{\thetable}{\Roman{table}}
\renewcommand{\thefigure}{\Roman{figure}}
\renewcommand\thesection{\Roman {section}}

\section*{Appendix}

\setcounter{section}{0}
\setcounter{figure}{0}
\setcounter{table}{0}
\providecommand{\titlevariable}{Biguide}

In the supplementary material, we provide additional visualizations and analyses, including: (1) a comparison of different target tracking strategies, (2) an evaluation of computation cost between end-to-end and modular baselines, and 
(3) a robustness study under perturbations in vehicle speed and target noise.

\subsection{Motion Prediction comparison}
We evaluate approaches for tracking both ego and target positions: (1) a segmentation-based method~\cite{yang2024e2e}, (2) a motion prediction model using a single historical frame, and (3) a motion prediction model leveraging multiple historical frames.

\vspace{-1mm}
\begin{table}[h]
\centering
\small
\renewcommand{\arraystretch}{1.5}
\captionsetup{font=sc}
\captionsetup{font={scriptsize, sc, stretch=1.3}, justification=centering, labelsep=newline}
\caption{Target tracking comparison using Average Displacement Error (ADE) and variance. Inference time is negligible. Note that MP. is short for motion prediction.}
\vspace{-1mm}
\resizebox{0.97\linewidth}{!}{%
\begin{tabular}{lcc}
\toprule
\rowcolor{blue!10} Method & FDE (m) $\downarrow$ & Variance $\downarrow$ \\
\midrule
Segmentation-based & 0.082 & 0.127 \\
Learnable MP.  (single-frame) & 0.403 & 0.020 \\
Dynamics (multi-frame) & \textbf{0.081} & \textbf{0.002} \\
\bottomrule
\end{tabular}%
}
\label{tab:tracking_fde_time}
\end{table}
\vspace{-1mm}

As shown in Tab.~\ref{tab:tracking_fde_time}, the segmentation-based method achieves moderate average error (FDE 0.082m) but is highly unstable, with variance reaching 0.127. This instability stems from its reliance on frame-by-frame segmentation: a single mis-segmented frame can lead to rapid divergence in subsequent motion and target predictions. Besides, the single-frame dynamics model cannot generalize well, resulting in a large FDE of 0.403m despite relatively low variance. In contrast, the multi-frame dynamics model achieves both the lowest error (0.081m) and minimal variance (0.002), as incorporating multiple historical frames smooths out noise and captures more reliable motion patterns, with negligible additional computational cost.

\subsection{Computation cost: E2E vs Modular}
\vspace{-2mm}
\begin{table}[h]
\centering
\small
\renewcommand{\arraystretch}{1.5}
\captionsetup{font=sc}
\captionsetup{font={scriptsize, sc, stretch=1.3}, justification=centering, labelsep=newline}
\caption{Computation cost comparison between E2EParking and modular baselines. Inference time refers to the average time to execute a full trajectory. For E2EP methods, GPU, CPU, and runtime metrics are averaged across all end-to-end variants for clarity.}
\vspace{-1mm}
\resizebox{0.97\linewidth}{!}{%
\begin{tabular}{lccc}
\toprule
\rowcolor{blue!10} Method & Memory Cost$\downarrow$ & CPU Cost$\downarrow$ & Inference Time (s)$\downarrow$ \\
\midrule
E2EParking (Ours) & 2182 MB & 6.1 GB & 118.25 \\
Modular: Hybrid A* & 2092 MB & 6.6 GB & 188.75 \\
\bottomrule
\end{tabular}%
}
\label{tab:comp_cost}
\end{table}
\vspace{-2mm}

Table~\ref{tab:comp_cost} reports the average memory usage, CPU load, and inference time for E2EParking and the modular Hybrid A* method. 
All end-to-end variants exhibit similar GPU, CPU, and runtime profiles, which are summarized as an average for clarity. 
Compared to the modular approach, E2EParking uses slightly more GPU memory but substantially less CPU, completing inference faster due to its single-pass end-to-end design. 
These results highlight the practical efficiency of end-to-end methods for real-time deployment.

\subsection{Resiliency to Speed and Target Noise}

We evaluate the robustness of \titlevariable~against the modular Hybrid A* method under perturbations in vehicle speed and target position, with noise applied as a standard deviation around the ground truth values. As shown in Tab.~\ref{tab:noise_robustness}, end-to-end methods are generally more resilient to both types of noise, particularly for target position, likely due to explicit target augmentation during training. In contrast, learnable dynamics models are less robust to speed noise, resulting in larger performance drops. Overall, while all methods exhibit proportionally higher sensitivity with increasing noise levels, \titlevariable~consistently maintains superior TSR stability, highlighting the benefits of its end-to-end design and target-aware training.

\vspace{-1mm}
\begin{table}[h]
\centering
\small
\renewcommand{\arraystretch}{1.5}
\captionsetup{font=sc}
\captionsetup{font={scriptsize, sc, stretch=1.3}, justification=centering, labelsep=newline}
\caption{Comparison of end-to-end and modular methods under vehicle speed and target position noise. TSR changes indicate the relative difference compared to the no-noise scenario.}
\vspace{-1mm}
\resizebox{0.92\linewidth}{!}{%
\begin{tabular}{lcccc}
\toprule
\rowcolor{blue!10} Method & Speed Noise ($\pm$\%) & TSR Change & Target Noise ($\pm$m) & TSR Change \\
\midrule
Modular Hybrid A* & 10 & -0.8\% & 0.2 & -0.3\% \\
Modular Hybrid A* & 30 & -4.9\% & 0.5 & -1.3\% \\
Biguide & 10 & -0.2\% & 0.2 & +0.2\% \\
Biguide & 30 & -3.5\% & 0.5 & -0.0\% \\
\bottomrule
\end{tabular}%
}
\label{tab:noise_robustness}
\end{table}

\clearpage

\end{document}